%% file: main.tex
\documentclass[letterpaper]{article} 
\usepackage{aaai2026}  
\usepackage{times}  
\usepackage{helvet}  
\usepackage{courier}  
\usepackage[hyphens]{url}  
\usepackage{graphicx} 
\usepackage{natbib}  
\usepackage{caption} 
\usepackage{algorithmicx}
\usepackage[ruled]{algorithm}
\usepackage{algpseudocode}
\algrenewcommand\algorithmicindent{0.75em}
\usepackage{amsmath}
\usepackage{amsfonts}
\usepackage{booktabs}
\usepackage{nicefrac}       
\usepackage{microtype}      
\usepackage{xcolor}         
\usepackage{pgfplots}
\usepgfplotslibrary{groupplots}
\pgfplotsset{compat=1.17}
\usepackage{newfloat}
\usepackage{listings}
\DeclareCaptionStyle{ruled}{labelfont=normalfont,labelsep=colon,strut=off} 
\floatstyle{ruled}
\newfloat{listing}{tb}{lst}{}
\floatname{listing}{Listing}
\title{FlashKAT: Understanding and Addressing Performance Bottlenecks in the Kolmogorov-Arnold Transformer}
\author{
    Matthew Raffel,
    Lizhong Chen
}
\affiliations{
    School of Electrical Engineering and Computer Science\\
    Oregon State University, USA\\
    \{raffelm, chenliz\}@oregonstate.edu
}

\begin{document}

\maketitle

\begin{abstract}
The Kolmogorov-Arnold Network (KAN) has been gaining popularity as an alternative to the multilayer perceptron (MLP) due to its greater expressiveness and interpretability. Even so, KAN suffers from training instability and being orders of magnitude slower due to its increased computational cost, limiting its applicability to large-scale tasks. Recently, the Kolmogorov-Arnold Transformer (KAT) has been proposed, achieving FLOPs comparable to traditional Transformer models with MLPs by leveraging Group-Rational KAN (GR-KAN). Unfortunately, despite the comparable FLOPs, our testing shows that KAT remains 123x slower during training, indicating that there are other performance bottlenecks beyond FLOPs. In this paper, we conduct a series of experiments to understand the root cause of the slowdown in KAT. We uncover that the slowdown can be isolated to memory stalls, linked more specifically to inefficient gradient accumulations in the backward pass of GR-KAN. To address this memory bottleneck, we propose FlashKAT, which minimizes accesses to slow memory and the usage of atomic adds through a restructured kernel. Evaluations show that FlashKAT achieves up to an 86.5x training speedup over state-of-the-art KAT while reducing rounding errors in gradient computation. 
\end{abstract}

\begin{links}
    \link{Code}{https://github.com/OSU-STARLAB/FlashKAT}
\end{links}

\section{Introduction}

The Kolmogorov-Arnold Network (KAN) has emerged as a popular alternative to the multilayer perceptron (MLP) \cite{liu2024kan}.  Rather than learning a fixed weight per edge, KAN instead learns a flexible nonlinearity. 
In doing so, KAN achieves greater expressivity and interpretability than MLP, which has led to impressive capabilities in scientific and equation modeling tasks \cite{li2025u, coffman2025matrixkan, liu2024kan, koenig2024kan, wang2024kolmogorov}.  However, even with the superior performance of KAN for these tasks, it has fallen short on larger-scale tasks in NLP and computer vision due to substantially increased parameters and computational cost, lack of GPU hardware optimization (e.g., for basis functions), and training instability \cite{yu2024kan, le2024exploring, VisionKAN2024}. For example, while computing an edge for an MLP only requires  2 FLOPS (1 for multiply and 1 for add), computing an edge for a KAN may require up to 204 FLOPs, due to the nonlinearity \cite{yang2024kolmogorov}. This shortcoming necessitates significant research to reduce the orders of magnitude performance gap between KANs and MLPs.

Recently, the Kolmogorov-Arnold Transformer (KAT) has overcome some of the obstacles that prevent KAN from being deployed to large-scale tasks \cite{yang2024kolmogorov}.  It achieves this by replacing each MLP in the Transformer with a Group-Rational KAN (GR-KAN), which augments the input to each linear layer with the safe Pade Activation Unit (PAU) \cite{molina2019pad} in a grouped fashion (more details in Section \ref{sec:background}).  In doing so, each GR-KAN learns a weighted summation of learned nonlinear functions. Unlike KAN, GR-KAN is more computationally efficient, reducing FLOPs to only slightly more than MLPs, and is designed to better map to GPUs. By modifying the Transformer with GR-KAN, the resulting KAT achieves impressive results on computer vision tasks.  

However, even with the comparable number of FLOPs (and similar \textit{expected} speed) offered by KAT, in our characterization, KAT is significantly slower than its MLP counterpart. For instance, our experiments reveal that a KAT (using GR-KAN) is 123x slower during training than an identically sized Vision Transformer (ViT, using MLP) \cite{dosovitskiy2020image} on ImageNet-1K \cite{russakovsky2015imagenet}.
This realization motivated an in-depth investigation to understand the  KAT's performance bottleneck. Unlike previous approaches focused primarily on FLOPs, we re-examine the performance issue from a memory-centric perspective, a novel angle not previously explored \cite{yang2024kolmogorov}, which uncovers that the slowdown is primarily due to a memory bottleneck rather than excessive computation. Specifically, we identify the inefficient gradient accumulation method for the PAU coefficients, involving atomic adds, as the key contributor \cite{molina2019pad}. This insight not only challenges assumptions that computation is the primary constraint in KAT but also offers a new direction for optimizing memory access patterns in similar architectures.  

To address the memory bottleneck, we propose FlashKAT, a \textit{substantially} faster KAT that solves GR-KAN's (and by extension PAU's) shortcomings by not only avoiding accumulating gradients with atomic adds but also the unnecessary accesses to high-bandwidth memory (HBM).  The method restructures data accesses and enables more effective use of shared memory for gradient accumulations. By implementing our method as a GPU kernel, FlashKAT achieves a training speedup of up to 86.5x compared with state-of-the-art KAT. Furthermore, we demonstrate that our new kernel reduces rounding errors in the PAU coefficient gradient computations by nearly an order of magnitude, thereby improving training stability.

As our method directly optimizes PAU, a learnable rational activation, the techniques and analysis also apply to alternative learnable rational activations, a complementary area of research to KAT \cite{molina2019pad, delfosse2020rationals, delfosse2021recurrent, biswas2021orthogonal}. Despite the clear need for efficient implementations, researchers in both the KAT and rational activation communities had not identified a way to accelerate these activations. FlashKAT provides this essential optimization, representing a nontrivial breakthrough that not only advances the state of the art but also inspires future research in the field.
 
The main contributions of this paper are:
\begin{enumerate}
   \item Conducting a detailed investigation to understand the computational inefficiencies of KAT, revealing that the primary performance bottleneck arises not from FLOPs, but from suboptimal memory accesses caused by inefficient gradient accumulation.
    \item Proposing FlashKAT, which addresses the identified memory bottleneck by leveraging a substantially more efficient gradient accumulation method to speed up training.
    \item Performing an extensive evaluation of FlashKAT, demonstrating significantly reduced training times and rounding errors from the proposed method.
\end{enumerate}

\section{Background and Related Works}
\label{sec:background}
\subsection{Kolmogorov-Arnold Network}
Recently, KAN has been popularized as an alternative to MLP \cite{liu2024kan}.  It is based on the Kolmogorov-Arnold Theorem \cite{kolmogorov1961representation}, which asserts that a multivariate continuous function on a bounded domain $f:[0,1]^n\rightarrow\mathbb{R}$ can be rewritten as a finite summation of  univariate continuous functions, $\phi_{q,p}:[0,1]\rightarrow \mathbb{R}$ and $\phi_q:\mathbb{R} \rightarrow \mathbb{R}$ as expressed in Equation \ref{eq:KAN}:

\begin{equation}
    \label{eq:KAN}
    f(x_1, ..., x_n) = \sum^{2n+1}_{q=1}\phi_q(\sum^n_{p=1}\phi_{q,p}(x_p)).
\end{equation}

Although Equation \ref{eq:KAN} presents a theoretical form of the Kolmogorov-Arnold Theorem, \cite{liu2024kan} demonstrated its effectiveness in a more generalized form by scaling the width and depth arbitrarily for a defined task. Suppose $n_l$ is the number of input nodes for layer $l$, where each input is represented by $x_{l, i}$.  Each edge connection to layer $l+1$ has a learned activation $\phi_{l,j,i}(\cdot)$, where $i\in[1,n_l]$ and $j\in[1,n_{l+1}]$. Then, we can represent each output node as

\begin{equation}
    \label{eq:KANGen}
    x_{l+1,j} = \sum^{n_l}_{i=1}\phi_{l,j,i}(x_{l,i}).
\end{equation}
In a vectorized format, we can represent Equation \ref{eq:KANGen} with
\begin{equation}
\label{eq:KANGenMat}
\mathbf{x}_{l+1} = \Phi \circ \mathbf{x}_l
=
\begin{bmatrix}
\phi_{l,1,1}(\,\cdot\,) & \cdots & \phi_{l,1,n_l}(\,\cdot\,) \\
\vdots               & \ddots & \vdots               \\
\phi_{l,n_{l+1},1}(\,\cdot\,) & \cdots & \phi_{l, n_{l+1},n_l}(\,\cdot\,)
\end{bmatrix}\mathbf{x}_l.
\end{equation}
Thus, an $L$ layer KAN can be represented as
\begin{equation}
    \label{eq:KANDeep}
    \text{KAN}(\mathbf{x})=(\Phi_L \circ \Phi_{L-1} \circ ... \circ \Phi_1)\mathbf{x}.
\end{equation}
The most widely adopted method for representing $\phi_{l,j,i}(\cdot)$ has been with B-splines \cite{liu2024kan}, and although B-spline-based KANs demonstrate promising results, they still have a host of shortcomings \cite{yang2024kolmogorov, yu2024kan}.  These shortcomings consist of (1) increased parameter and computational cost compared to MLP (e.g., increased FLOPs as summarized in Table \ref{tab:Flopequations}), (2) B-splines not being GPU optimizable (e.g., requiring a recursive algorithm), and (3) training instability in large networks \cite{yang2024kolmogorov}.  Such shortcomings limit KAN's compatibility with the Transformer \cite{vaswani2017attention}, restricting its applicability to large-scale tasks like computer vision.

\subsection{Kolmogorov-Arnold Transformer}
Unlike its predecessors, littered with shortcomings, KAT succeeds in merging KAN into a Transformer.  It does so by developing a new KAN variation called GR-KAN, which (1) shares coefficients across edge groups, reducing FLOPs, (2) leverages safe PAU activation functions, eliminating recursion, and (3) applies a variance-preserving initialization, increasing training stability \cite{yang2024kolmogorov}. In KAT, each MLP is replaced with a GR-KAN, causing each MLP weight to be augmented with a nonlinear learned activation function (i.e., a nonlinear edge in KAN). 

\begin{table*}[ht]
\centering
\small
\begin{tabular}{lll}
\toprule
Name & Number of parameters & FLOPs \\
\midrule
MLP (ViT) &
$d_{\mathrm{in}}\times d_{\mathrm{out}}$ &
$\mathrm{FuncFLOPs}\times d_{\mathrm{out}} + 2 \times d_{\mathrm{in}}\times d_{\mathrm{out}}$ \\[1ex]

KAN &
$d_{\mathrm{in}}\times d_{\mathrm{out}}\times(G + K + 3)$ &
\begin{tabular}{@{}l@{}}
$\mathrm{FuncFLOPs}\times d_{\mathrm{in}} 
  + d_{\mathrm{in}}\times d_{\mathrm{out}}\times\bigl[9K\times (G + 1.5K)+2G -2.5K+3\,\bigr]$
\end{tabular} \\[2ex]

GR‐KAN (KAT) &
$d_{\mathrm{in}}\times d_{\mathrm{out}} + (m + n\times g + 1)$ &
$(2m + 2n + 3) \times d_{\mathrm{in}} + 2\times d_{\mathrm{in}}\times d_{\mathrm{out}}$ \\
\bottomrule
\end{tabular}
\caption{Comparison of parameter counts and FLOPs for different layers. “FuncFLOPs” denotes the activation FLOPs. In KAN, $K$ is the spline order and $G$ the number of intervals; in GR‐KAN, $m,n$ are the polynomial degrees and $g$ the number of groups.}
\label{tab:Flopequations}
\end{table*}

First, rather than learning $d_{in}\cdot d_{out}$ activations, $\phi_{l,j,i}(\cdot)$, like KAN, GR-KAN groups inputs and only learns $\lfloor d_{in}/d_g\rfloor$ activations, $F(\cdot)$, where $d_g$ is the number of activations that share the same set of learned coefficients in a group.  Each output $x_{l+1,j}$ is then a weighted summation of $F_{\lfloor i/d_g \rfloor}(x_{l,i})$ using learnable weights, $\mathbf{W}\in \mathbb{R}^{d_{out}\times d_{in}}$.  GR-KAN is expressed in Equation \ref{eq:GR_KAN}:  

\begin{equation}
\label{eq:GR_KAN}
    \begin{split}
    &\text{GR-KAN}(\mathbf{x}) = \, \mathbf{W}\mathbf{F}(\mathbf{x})\\
    & =\begin{bmatrix}
    w_{11} & \cdots & w_{1,d_{in}} \\
    \vdots & \ddots & \vdots \\
    w_{d_{out},1} & \cdots & w_{d_{out}, d_{in}}
    \end{bmatrix} 
    \begin{bmatrix}
    F_{\lfloor1/d_g\rfloor}(x_1) \\
    \vdots \\
    F_{\lfloor d_{in}/d_g\rfloor}(x_{d_{in}})
    \end{bmatrix}
    \end{split}.
\end{equation}

Second, each GR-KAN activation is a safe PAU \cite{molina2019pad}, which we will refer to as a \textit{group-wise rational function}, composed of learnable coefficients $a_i$ and $b_j$ as follows:
\begin{equation}
    \label{eq:pade}
    F(x) = \frac{P(x)}{Q(x)}=\frac{a_0+a_1x+a_2x^2+...+a_mx^m}{1+|b_1x+b_2x^2+...+b_nx^n|}.
\end{equation}

Third, a variance-preserving initialization is applied for stable learning \cite{yang2024kolmogorov}.  The procedure for applying variance preserving weights requires (1) initializing the coefficients of $F(\cdot)$ to mimic a known activation function (ie. Swish) and (2) initializing the weights in $\mathbf{W}$ according to $\mathcal{N}(0, \frac{\alpha}{d_{in}})$ whereby $\alpha=\frac{\mathbb{E}[\mathbf{F(x)}^2]}{Var[\mathbf{x}]}$ assuming  $\mathbf{x}\sim \mathcal{N}(0,1)$.

Although KAT is effective at vision tasks and is more computationally efficient than a version of ViT with KAN (a result of overcoming shortcomings (1), (2), and (3)), they still far short in terms of training speed compared to an identical ViT using MLP, a shortcoming which has slowed further research. We demonstrate such results in the next Section.

\section{Beyond FLOPs: Understanding Performance Bottlenecks of KAT}

\subsection{Insight 1: KAT Is Significantly Slower Than ViT During Training}
\label{sec:Flop}
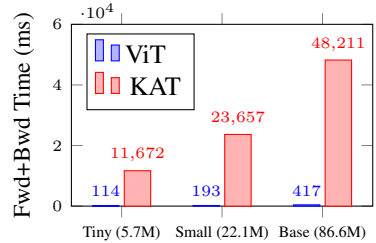
\begin{figure}[h]  
  \centering
  \begin{tikzpicture}
    \begin{axis}[
      ybar,
      bar width=15pt,
      width=0.5\textwidth, height=4cm,  
      symbolic x coords={Tiny (5.7M), Small (22.1M), Base (86.6M)},
      xtick=data,
      ylabel=Fwd+Bwd Time (ms),
      label style={font=\small},
      tick label style={font=\small},
      enlarge x limits=0.25,
      ymin=0,
      ymax=60000,
      legend style={
        at={(0.05,0.95)},
        anchor=north west,
        /tikz/every even column/.append style={column sep=4pt}
      },
      legend cell align=left,
      nodes near coords,
      every node near coord/.append style={font=\tiny},
    ]
      \addplot coordinates {(Tiny (5.7M),114) (Small (22.1M), 193) (Base (86.6M),417)};
      \addplot coordinates {(Tiny (5.7M),11672) (Small (22.1M),23657) (Base (86.6M),48211)};
      \legend{ViT,\,KAT}
    \end{axis}
  \end{tikzpicture}
  \caption{Comparison of training time (Fwd+Bwd) for ViT and KAT.}
  \label{fig:train-times}
\end{figure}

Although KAT presents a promising alternative to ViT, it still lags behind ViT in training time. We illustrate this by comparing the combined forward and backward times for KAT-T (5.7M parameters), KAT-S (22.1M parameters), and KAT-B (86.6M parameters) against identically sized ViT-T, ViT-S, and ViT-B models, using the ImageNet-1k dataset for training \cite{russakovsky2015imagenet}. For the experiment, each KAT group-wise rational function contains 8 groups comprising of 6 numerator and 4 denominator coefficients, and all measurements are taken using an H200 GPU. The results, depicted in Figure \ref{fig:train-times}, show the average time of 100 forward and backward passes, following 5 warmup passes.  The data reveals that, for these forward/backward passes, KAT-T is 102x slower than ViT-T, KAT-S is 123x slower than ViT-S, and KAT-B is 116x slower than ViT-B.

\subsection{Insight 2: FLOPs Is Not the Performance Bottleneck}

\label{sec:scalingFlops}
It is natural to assume that such a drop-off in training speed between ViT and KAT results from the difference in FLOPs, as KAT has more FLOPs than ViT. However, by comparing the estimated FLOPs for GR-KAN and MLP (Table \ref{tab:Flopequations}), the only component in ViT and KAT where they differ, we can dismiss this assumption.

As can be seen, between the FLOPs expressions of ViT and KAT, the main difference resides in the MLP activation requiring $\text{FuncFLOPs} \times d_{out}$ FLOPs and the GR-KAN learnable activation requiring $(2m+2n+3) \times d_{in}$ FLOPs, both of which are negligible compared with the $2 \times d_{in} \times d_{out}$ term. This suggests that the slight increase in FLOPs in KAT may not be the cause of the slowdown. To further verify this, we conduct an experiment that artificially increases the FLOPs inside the group-wise rational function of KAT by placing an additional loop around the floating-point computations. This would examine whether FLOPs bottleneck the performance. We measure performance metrics using Nvidia Nsight Compute\footnote{The documentation for Nvidia Nsight Compute can be found at https://docs.nvidia.com/nsight-compute/} on an RTX 4060 Ti for the forward and backward pass.  The dimension configuration is: input tensor, $\mathbf{X}\in\mathbb{R}^{1024\times 197\times768}$, numerator coefficients, $\mathbf{A}\in \mathbb{R}^{8\times6}$, denominator coefficients, $\mathbf{B}\in \mathbb{R}^{8\times4}$, and the upstream gradient tensor, $\mathbf{dO}\in\mathbb{R}^{1024\times 197\times768}$. We report these results in Table \ref{tab:pass_performance}.

\begin{table*}[ht]
\centering
\small
\begin{tabular}{cccccccc}
\toprule
\multicolumn{8}{c}{\bfseries Forward Pass} \\
\midrule
Loops
& FLOPs
& Cycles
& Time
& SM Thp. (\%)
& L1 Thp. (\%)
& L2 Thp. (\%)
& HBM Thp. (\%) \\
\midrule
1 & 2.9T & 11.3M & 4.89 ms & 29.14 & 28.25 & 18.11 & 89.40 \\
2 &  5.9T & 11.3M & 4.91 ms & 30.34 & 28.46 & 18.08 & 89.25 \\ 
4 & 11.8T & 11.3M & 4.89 ms & 41.02 & 29.08 & 19.78 & 89.37 \\
8 & 23.6T & 11.3M & 4.91 ms & 62.20 & 28.00 & 18.40 & 89.34 \\
\midrule
\multicolumn{8}{c}{\bfseries Backward Pass} \\
\midrule
Loops &
FLOPs
& Cycles
& Time
& SM Thp. (\%)
& L1 Thp. (\%)
& L2 Thp. (\%)
& HBM Thp. (\%) \\
\midrule
1 & 11.2T & 2.4T & 1.03 s & 1.97 & 4.38 & 5.24 & 1.01 \\
2 & 22.3T & 2.4T & 1.03 s & 1.97 & 4.38 & 5.22 & 1.01 \\
4 & 44.6T & 2.4T & 1.03 s & 1.97 & 4.38 & 5.22 & 1.01 \\
8 & 89.2T & 2.4T & 1.03 s & 1.97 & 4.38 & 5.22 & 1.01 \\
\bottomrule
\end{tabular}
\caption{Performance breakdown for forward and backward passes of the group-wise rational function. ``Loops'' is the total instruction loops; ``Cycles'' is the cycle count; ``Time'' is wall-clock time; ``FLOPs'' is the floating point operation count; ``SM Thp.'', ``L1 Thp.'', ``L2 Thp.'', and ``HBM Thp.'' report the percentage of the device’s peak arithmetic and memory throughput.}
\label{tab:pass_performance}
\end{table*}

From Table \ref{tab:pass_performance}, we can immediately see that, even if the FLOPs scale to 8x of the original amount, the overall required ``Cycles'' or ``Time'' to complete the computation does not scale in either the forward or backward pass.  Such results clearly demonstrate that the forward and backward pass kernels in KAT are not compute-bound, and we must look beyond FLOPs to identify performance bottlenecks.

\subsection{Insight 3: The Backward Pass Dominates Execution Time}

Results in Table \ref{tab:pass_performance} also demonstrate that the group-wise rational function dedicates a far greater amount of compute time to the backward pass than the forward pass.  For instance, the forward pass takes 4.96 ms, whereas the backward pass requires 1.03 s, which is 207.7x as long.  Therefore, to reduce KAT's training time, performance issues in the backward pass must be identified and addressed. The backward pass can be broken down into three components: (1) loading values to compute the gradients, (2) computing the gradients, and (3) storing the gradients.  Since we know the backward pass is not compute-bound from Insight 2, this means that (1) and (3) are to blame for the slowdown, both of which are memory issues.        

\subsection{Insight 4: Memory (and Memory Stall in Particular) Is the Culprit}

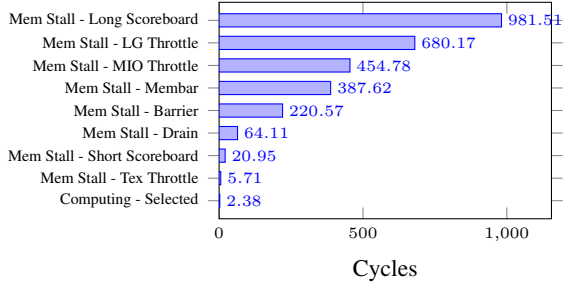
\begin{figure}[h]
  \centering
  \begin{tikzpicture}
    \begin{axis}[
      xbar,
      bar width=5pt,
      width=6.5cm,
      height=5cm,           
      y=0.35cm,
      xmin=0,
      xmax=1050,
      xlabel={Cycles},
      label style={font=\small},
      tick label style={font=\small},
      symbolic y coords={
        MS - Long Scoreboard,
        MS - LG Throttle,
        MS - MIO Throttle,
        MS - Membar,
        MS - Barrier,
        MS - Drain,
        MS - Short Scoreboard,
        MS - Tex Throttle,
        Computing - Selected
      },
      ytick=data,
      y dir=reverse,         
      enlarge y limits=0.1,    
      enlarge x limits={upper=0},
      nodes near coords,
      every node near coord/.append style={
        font=\tiny,
        anchor=west,
        xshift=-1pt
      },
    ]
    \addplot coordinates {
      (981.51,MS - Long Scoreboard)
      (680.17,MS - LG Throttle)
      (454.78,MS - MIO Throttle)
      (387.62,MS - Membar)
      (220.57,MS - Barrier)
      (64.11,MS - Drain)
      (20.95,MS - Short Scoreboard)
      (5.71,MS - Tex Throttle)
      (2.38,Computing - Selected)
    };
    \end{axis}
  \end{tikzpicture}
  \caption{The warp‐state statistics for the KAT group-wise rational function backward pass.  ``Computing - Selected'' is where the warp does useful computation; all the others are memory stall (MS) states.}
  \label{fig:warp‐cycles‐named}
\end{figure}
Focusing on the memory side for backward passes, we note that Table \ref{tab:pass_performance} reports very low L1, L2, and HBM throughput. At first glance, this is counterintuitive because a memory-bound process would typically imply a saturated memory bandwidth. However, further reasoning about the underlying hardware can provide insights. 
A GPU consists of multiple streaming multiprocessors (SMs). Typically, 32 threads are grouped in a \textit{warp} and executed synchronously. Multiple warps can be scheduled for execution on an SM in a context-switching fashion. When one warp is waiting for a memory transfer, the SM can switch to doing computation for a different ``ready'' warp (i.e., not waiting for memory), thereby hiding memory latency. However, suppose there are not enough ready warps. Then, the warp scheduler cannot entirely hide long latencies or high memory contention, like with atomic operations that create sequential memory accesses. If an SM does not make adequate use of the memory hierarchy consisting of registers, shared memory, and global memory (a mapped memory composed of L1 cache, L2 cache, and HBM), the latency for each transfer becomes even more severe.  For reference,  on an A100, shared memory, L1, L2, and HBM have latencies of 29.0, 37.9, 261.5, and 466.3 clock cycles, respectively  \cite{luo2024benchmarking}. Assuming that the group-wise rational function evaluation fails to hide latency, it makes sense that we would witness low memory utilization yet still experience a memory bottleneck. To validate this assumption, we examine the warp state statistics for the group-wise rational function backward pass using Nvidia Nsight Compute with the same configuration as before. These statistics report the average number of cycles each warp remains in a given state for each instruction the kernel executes.  We report the warp state statistics in Figure \ref{fig:warp‐cycles‐named}.

Figure \ref{fig:warp‐cycles‐named} demonstrates that each warp spends the majority of its time waiting for various memory stalls to conclude.  Each reported stall corresponds to a memory transfer, excluding the ``Selected'' state, in which the warp is selected to issue an instruction (i.e., compute).  For instance, the ``Stall Long Scoreboard'', a stall on a memory transfer from global memory, accounts for 412x the time spent in the ``Selected'' state.  As such, it becomes clear that the group-wise rational function slowdown is primarily due to memory transfer stalls. This calls for a new method that is aware of the memory hierarchy and avoids memory operations that create contention, limiting the warp scheduler's latency-hiding capabilities.

\section{FlashKAT}
In the previous section, we identified that KAT has a memory bottleneck in the backward pass of the group-wise rational function, a finding that prior approaches had not recognized \cite{yang2024kolmogorov,molina2019pad,delfosse2020rationals,delfosse2021recurrent}. In this section, we first outline the computations required for this operation, then provide an analysis of the standard group-wise rational function used in KAT to pin down the source of the bottleneck, and finally present FlashKAT, our proposed Kolmogorov-Arnold Transformer that builds on an optimized group-wise rational function that efficiently leverages gradient accumulation and memory hierarchy to minimize global memory accesses.

\subsection{Gradient Computations}
Since group-wise rational function, expressed in Equation \ref{eq:pade} as $F(\cdot)$, is an element-wise operation composed of learnable parameters $a_i$ and $b_j$ for the numerator and the denominator, the backward pass requires us to compute the gradients $\frac{\partial F}{\partial a_i}$, $\frac{\partial F}{\partial b_j}$, and $\frac{\partial F}{\partial x}$. Suppose that $A(x)=b_1x+...,+b_nx^n$, $\frac{\partial P(x)}{\partial x}=a_1+2a_2x+...+ma_mx^{m-1}$, and $\frac{\partial Q(x)}{\partial x}=\frac{\partial (1+|A(x)|)}{\partial A(x)}\cdot\frac{\partial A(x)}{\partial x}=\frac{A(x)}{|A(x)|}(b_1+2b_2x+...+nb_nx^{n-1})$ then the gradients for each are

\begin{align}
\label{eq:grad_a}
&\frac{\partial F}{\partial a_i} = \frac{x^i}{Q(x)}, \\
\label{eq:grad_b}
&\frac{\partial F}{\partial b_j} = -x^j\frac{ A(x)}{|A(x)|} \cdot \frac{P(x)}{Q(x)^2},\\
\label{eq:grad_x}
&\frac{\partial F}{\partial x} = \frac{\partial P(x)}{\partial x} \cdot \frac{1}{Q(x)} - \frac{\partial Q(x)}{\partial x} \cdot \frac{P(x)}{Q^2(x)}.
\end{align}

Equations \ref{eq:grad_a}, \ref{eq:grad_b}, and \ref{eq:grad_x} provides the local element-wise gradients; however, since each batch of size $B$, each element in the sequence of size $N$, and each group of size $d_g$ share an identical set of coefficients, the contributions of each must be accumulated on $\frac{\partial F}{\partial a_i}$ and $\frac{\partial F}{\partial b_j}$.  Let us denote $\frac{\partial F}{\partial a_{g,i}}$ and $\frac{\partial F}{\partial b_{g,j}}$ as the gradient for coefficient $i$ and $j$ of group $g$ and $\frac{\partial \mathcal{L}}{\partial F}$ the upstream gradient.  Then we can express this accumulation in Equations \ref{eq:accum} and \ref{eq:accum2}:
\begin{align}
\label{eq:accum}
    \frac{\partial \mathcal{L}}{\partial a_{g,i}}=\sum^B_{i'=1}\sum^N_{j'=1}\sum^{d_g}_{k'=1} \frac{\partial \mathcal{L}}{\partial F^{i',j'}_{g\cdot d_g+k'}}\cdot\frac{\partial F^{i',j'}_{g\cdot d_g+k'}}{\partial a_{g,i}},\\  
\label{eq:accum2}
    \frac{\partial \mathcal{L}}{\partial b_{g,j}}=\sum^B_{i'=1}\sum^N_{j'=1}\sum^{d_g}_{k'=1}\frac{\partial \mathcal{L}}{\partial F^{i',j'}_{g\cdot d_g+k'}}\cdot \frac{\partial F^{i',j'}_{g\cdot d_g+k'}}{\partial b_{g,j}}.
\end{align}
There are multiple different methods for accumulating these gradients in a kernel, and identifying the optimal accumulation procedure is crucial for a fast training algorithm.  We will now demonstrate where the standard group-wise rational function method falls short in this regard. 

\subsection{Analysis of KAT Group-Wise Rational Function}
Provided an input $\mathbf{X}\in \mathbb{R}^{B\times N\times d}$, an upstream gradient $\mathbf{dO}\in \mathbb{R}^{B\times N\times d}$ and coefficients $\mathbf{A} \in \mathbb{R}^{n_g\times m+1}$ and $\mathbf{B} \in \mathbb{R}^{n_g\times n}$, we want to compute the gradients $\mathbf{dA}\in \mathbb{R}^{n_g\times m+1}$ ($\frac{\partial \mathcal{L}}{\partial \mathbf{A}}$), $\mathbf{dB}\in \mathbb{R}^{n_g\times n}$ ($\frac{\partial \mathcal{L}}{\partial \mathbf{B}}$), and $\mathbf{dX}\in \mathbb{R}^{B\times N\times d}$ ($\frac{\partial \mathcal{L}}{\partial \mathbf{X}}$).  In Algorithm \ref{alg:F_slow_alg}, we provide the method for computing each of these gradients in KAT.  The algorithm is structured sequentially, and all on-chip computation is performed within subroutine calls, with the upstream gradients applied to Equations \ref{eq:grad_a}, \ref{eq:grad_b}, and \ref{eq:grad_x} for readability.
\begin{algorithm}[t]
  \small
  \caption{The KAT group-wise rational function backward pass.}
    \label{alg:F_slow_alg}
  \begin{algorithmic}[1]
    \Require 
      $\mathbf{X}, \mathbf{dO} \in \mathbb{R}^{B\times N\times d}$,\;
      $\mathbf{A} \in \mathbb{R}^{n_g\times m+1}$,  \;
      $\mathbf{B} \in \mathbb{R}^{n_g\times n}$ in global memory.
      
    \State Set block size to $S_{\text{block}}$.
    \State Divide $\mathbf{X}$ and $\textbf{dO}$ into $T=\lceil (B\cdot N\cdot d/S_\text{block}\rceil$ blocks $\mathbf{X}_1, ..., \mathbf{X}_T$ and $\mathbf{dO}_1, ..., \mathbf{dO}_T$ each of size $S_{\text{block}}$.  
    \State Initialize $\textbf{dA}\gets 0\in\mathbb{R}^{n_g\times m+1}$, $\mathbf{dB}\gets 0\in\mathbb{R}^{n_g\times n}$, and $\textbf{dX}\gets 0\in\mathbb{R}^{B\times N\times d}$.  

    \For{$1\leq i \leq T$}
        \State Load $\mathbf{X}_i$ and $\mathbf{dO}_i$ from global memory
        \For{$1\leq j \leq S_{\text{block}}$}
            \State $k \gets \lfloor(((i-1)\cdot S_\text{block}+j)\mod d)/d_g\rfloor$
            \State Load $\mathbf{A}_{k, :}$ and $\mathbf{B}_{k,:}$ from global memory
            \State $\mathbf{dX}_{i,j} \gets \textbf{Comp\_dX}(\mathbf{A}_{k, :}, \mathbf{B}_{k,:}, \mathbf{X}_{i,j}, \mathbf{dO}_{i,j})$
            \State $\mathbf{dA}'_{k,:} \gets \textbf{Comp\_dA}(\mathbf{A}_{k,:}, \mathbf{B}_{k,:}, \mathbf{X}_{i,j}, \mathbf{dO}_{i,j})$
            \State $\mathbf{dB}'_{k,:} \gets \textbf{Comp\_dB}(\mathbf{A}_{k, :}, \mathbf{B}_{k,:}, \mathbf{X}_{i,j}, \mathbf{dO}_{i,j})$
            \State Load $\mathbf{dA}_{k,:}$ from global memory, $\mathbf{dA}_{k,:} \gets \mathbf{dA}_{k,:} + \mathbf{dA}'_{k,:}$, write back $\mathbf{dA}_{k,:}$ to global memory \Comment{atomic add}
            \State Load $\mathbf{dB}_{k,:}$ from global memory, $\mathbf{dB}_{k,:} \gets \mathbf{dB}_{k,:} + \mathbf{dB}'_{k,:}$, write back $\mathbf{dB}_{k,:}$ to global memory \Comment{atomic add}
        \EndFor
        \State Write back $\mathbf{dX}_i$ to global memory
    \EndFor

    \State \Return $\mathbf{dA},\mathbf{dB},\mathbf{dX}$
  \end{algorithmic}
\end{algorithm}

At a high level, Algorithm \ref{alg:F_slow_alg} iterates over the $T$ blocks in the grid, where each block has $S_\text{block}$ threads that compute the associated gradients for $\mathbf{dA}$, $\mathbf{dB}$, $\mathbf{dX}$.  In the case of $\mathbf{dA}$ and $\mathbf{dB}$, for each of these computations, the thread performs an atomic add, which requires reading from global memory, accumulating the value, and then storing the result back to global memory as an indivisible unit of work.  However, since each of these threads must perform an atomic add for each of the $m+n+1$ coefficients, the atomic adds must occur sequentially.  As such, for each of these atomic adds, a stall occurs.  Since multiple threads within the same block and across different blocks may write to the same location, there will be severe resource contention.  Such resource contention inevitably leads to high latency and stalls without utilizing the available bandwidth, as we demonstrated in Section \ref{sec:Flop}.  

Let us analyze the total number of global memory accesses incurred by Algorithm \ref{alg:F_slow_alg}. We can see that each $\mathbf{X}_i$ and $\mathbf{dO}_i$ requires $S_\text{block}$ reads from global memory.  Then writing $\mathbf{dX}_i$ back to global memory requires $S_\text{block}$ writes to global memory.  Since these 3 memory actions occur $T$ times they require $3TS_\text{block}=3\frac{B\cdot N \cdot d}{S_\text{block}}\cdot S_\text{block}=3\cdot B\cdot N \cdot d$ global memory accesses.  Regarding the coefficients, each element in the output $dX$ requires loading $m+n+1$ coefficients from global memory.  It also requires the aforementioned atomic add, which includes an additional $m+n+1$ reads and writes to global memory.  Thus, since there are $T$ blocks of $S_\text{block}$ threads, there are $TS_\text{block}\cdot3(m+n+1)=3(m+n+1)\cdot B \cdot N \cdot d$ accesses.  As a result, there are $3(m+n+2)\cdot B \cdot N \cdot d$ global memory accesses in total.  To reiterate, memory accesses for atomic add operations for coefficient accumulations are much worse due to additional contention from other threads.

\subsection{FlashKAT Group-Wise Rational Function}
We now present an optimized method for computing the backward pass of the group-wise rational function in Algorithm \ref{alg:fast_F_alg}.  To reduce global memory accesses, we restructure the grid from 1D to 2D.  Such a restructuring mirrors the format of Equation \ref{eq:accum}, facilitating a more straightforward mapping of it to the algorithm.  The first dimension of our reconstructed grid is allocated for resolving the $B$ and $N$ summations, whereas the second dimension is responsible for the $d_g$  summation.  The size of this first dimension in the grid is $B\cdot N/S_\text{block}$, whereas the size of the second dimension is $n_g$ (the total number of groups in GR-KAN).

\begin{algorithm}[t]
  \small
  \caption{The FlashKAT  group-wise rational function backward pass.}
  \label{alg:fast_F_alg}
  \begin{algorithmic}[1]
    \Require 
      $\mathbf{X},\mathbf{dO} \in \mathbb{R}^{B\times N\times d}$,\;
      $\mathbf{A} \in \mathbb{R}^{n_g\times m+1}$,  \;
      $\mathbf{B} \in \mathbb{R}^{n_g\times n}$ in global memory.
      
    \State Set block sizes to $S_\text{block}$ and $d_g=d/n_g$.
    \State Divide $\mathbf{X}$ and $\mathbf{dO}$ into $\lceil (B\cdot N\cdot d/(S_\text{block}\cdot d_g)\rceil$ blocks $\mathbf{X}_{i,j}\in\mathbb{R}^{S_\text{block}\times d_g}$ and $\mathbf{dO}_{i,j}\in\mathbb{R}^{S_\text{block}\times d_g}$ of a $T\times n_g$ grid where $T=\lceil (B\cdot N/S_\text{block}\rceil$. 
    \State Divide $\mathbf{A}$ and $\mathbf{B}$ into $n_g$ blocks $\mathbf{A}_j\in \mathbb{R}^{m+1}$ and $\mathbf{B}_j\in \mathbb{R}^{n}$.
    \State Initialize $\mathbf{dA}\gets 0\in\mathbb{R}^{n_g\times m+1}$, $\mathbf{dB}\gets 0\in\mathbb{R}^{n_g\times n}$, and $\mathbf{dX}\gets 0\in\mathbb{R}^{B\times N\times d}$.  

    \For{$1\leq i \leq T$}
    \State  \hspace{-1em} \textbf{for} $1\leq j \leq n_g$ \textbf{do}
            \State Load $\mathbf{A}_j$, $\mathbf{B}_j$, $\mathbf{X}_{i,j}$, and $\mathbf{dO}_{i,j}$ from global memory 
            \State Initialize $\mathbf{dA}_j' \gets 0 $ and $\mathbf{dB}_j' \gets 0 $
            \For{$1\leq k \leq S_{\text{block}}$}
            \State \hspace{-1em} \textbf{for} $1\leq l \leq d_g$ \textbf{do}
                    \State $\mathbf{dX}_{i,j,k,l} \gets \textbf{Comp\_dX}(\mathbf{A}_{j}, \mathbf{B}_{j}, \mathbf{X}_{i,j,k,l}, \mathbf{dO}_{i,j,k,l})$
                    \State $\mathbf{dA}'_j \gets \mathbf{dA}'_j + \textbf{Comp\_dA}(\mathbf{A}_{j}, \mathbf{B}_{j}, \mathbf{X}_{i,j,k,l}, \mathbf{dO}_{i,j,k,l})$
                    \State $\mathbf{dB}'_j \gets \mathbf{dB}'_j + \textbf{Comp\_dB}(\mathbf{A}_{j}, \mathbf{B}_{j}, \mathbf{X}_{i,j,k,l}, \mathbf{dO}_{i,j,k,l})$
            \State \hspace{-1em} \textbf{end for}
            \EndFor
            \State Load $\mathbf{dA}_j$ from global memory, $\mathbf{dA}_j \gets \mathbf{dA}_j + \mathbf{dA}'_j$, write back $\mathbf{dA}_j$ to global memory \Comment{atomic add}
            \State Load $\mathbf{dB}_j$ from global memory, $\mathbf{dB}_j \gets \mathbf{dB}_j + \mathbf{dB}'_j$, write back $\mathbf{dB}_j$ to global memory \Comment{atomic add}
            \State Write back $\mathbf{dX}_{i,j}$ to global memory 
    \State \hspace{-1em} \textbf{end for}
    \EndFor

    \State \Return $\mathbf{dA},\mathbf{dB},\mathbf{dX}$
  \end{algorithmic}
\end{algorithm}

Algorithm \ref{alg:fast_F_alg} does not change the memory accesses for reads and writes for $\mathbf{dX}, \mathbf{X}$ and $\mathbf{dO}$; however, it heavily influences them for $\mathbf{A}$, $\mathbf{B}$, $\mathbf{dA}$, and $\mathbf{dB}$. At a high level, we can see that Algorithm \ref{alg:fast_F_alg} iterates over the grid of $T\cdot n_g$ blocks.  For each of these blocks, it loads a single group of coefficients for $\mathbf{A}_j$ and $\mathbf{B}_j$, which it then reuses for all computations in the block.  As a result, rather than each thread performing an atomic add for each value in $\mathbf{dA}_j$ and $\mathbf{dB}_j$, the algorithm can instead accumulate all the contributions for $\mathbf{dA}_j$ and $\mathbf{dB}_j$ in the block before doing an atomic add.  As a result, Algorithm \ref{alg:fast_F_alg} reduces the total amount of memory contention and global memory accesses.  

We now analyze the total global memory accesses for Algorithm \ref{alg:fast_F_alg}.  Each block first loads one set of coefficients from the global memory associated with the group to which the block is assigned.  Since there is only 1 set of coefficients per block, this operation requires only $m+n+1$ accesses.  Then the algorithm loads the associated $\mathbf{X}_{i,j}$ and $\mathbf{dO}_{i,j}$ from global memory, each of size $S_\text{block}\cdot d_g$. After computing $\mathbf{dX}_{i,j}$, its $S_\text{block}\cdot d_g$ values are written back to global memory.  Then, an atomic add loads $m+n+1$ values from $\mathbf{dA}_j$ and $\mathbf{dB_j}$, accumulates them with $\mathbf{dA}'_j$ and $\mathbf{dB}'_j$, and writes them back to global memory.  Since all these operations occur across the $T\cdot n_g$ blocks, there are $T\cdot n_g(3S_\text{block}\cdot d_g + 3(m+n+1))=3(\frac{m+n+1}{S_\text{block}\cdot d_g}+1)\cdot B\cdot N\cdot d$ global memory access.  Compared to Algorithm \ref{alg:F_slow_alg}, we can see that Algorithm \ref{alg:fast_F_alg} reduces the number of global memory accesses and atomic adds by a factor of $\frac{1}{S_\text{block}\cdot d_g}$.

\section{Results}
\label{sec:results}
\subsection{Revisiting Speed, Utilization and Stalls}
Using our experimental configurations from Section \ref{sec:Flop}, we verify the efficacy of Algorithm \ref{alg:fast_F_alg} compared to Algorithm \ref{alg:F_slow_alg}. We implement Algorithm \ref{alg:fast_F_alg} in Triton \cite{10.1145/3315508.3329973} and run our experiments on a 4060 Ti. 
First, we compare the backward pass of the FlashKAT group-wise rational function with that of the KAT group-wise rational function by measuring cycles, time, and throughput utilization for SMs and memory. As shown in Table \ref{tab:pass_eff_performance}, the FlashKAT group-wise rational function achieves a speedup of 140.5x over the KAT group-wise rational function, with improvements across SM, cache, and memory utilization, reducing backward pass time.  

\begin{table*}[t]
\centering
\small
\begin{tabular}{cccccccc}
\toprule
Model
& Cycles
& Time
& SM Thp. (\%)
& L1 Thp. (\%)
& L2 Thp. (\%)
& HBM Thp. (\%) \\
\midrule
KAT & 2.4T & 1.03 s & 1.97 & 4.38 & 5.24 & 1.01 \\
FlashKAT & 16.9M & 7.33 ms & 32.24 & 34.14 & 44.76 & 92.05 \\
\bottomrule
\end{tabular}
\caption{A comparison between the group-wise rational function backward pass for KAT and FlashKAT using FLOPs, cycles, time, SM, and memory throughput.}
\label{tab:pass_eff_performance}
\end{table*}

\begin{figure}[h]
  \centering
  \begin{tikzpicture}
    \begin{axis}[
      xbar,
      bar width=5pt,
      width=6.5cm,
      xmin=0,
      xmax=2.5,
      height=2cm,
      y=0.35cm,
      xlabel={Cycles},
      label style={font=\small},
      tick label style={font=\small},
      symbolic y coords={
        MS - Long Scoreboard,
        Computing - Selected
      },
      ytick=data,
      y dir=reverse,                      
      enlarge y limits=1,  
      enlarge x limits={upper=0},
      nodes near coords,
      every node near coord/.append style={
        font=\tiny,
        anchor=west,    
        xshift=-1pt      
      },
    ]
    \addplot coordinates {
      (2.31,MS - Long Scoreboard)
      (1.00,Computing - Selected)
    };
    \end{axis}
  \end{tikzpicture}
  \caption{The warp‐state statistics for the FlashKAT group-wise rational function backward pass.  ``Computing - Selected'' is where the warp does useful computation; all the others are memory stall (MS) states.}
  \label{fig:warp‐cycles‐eff}
\end{figure}
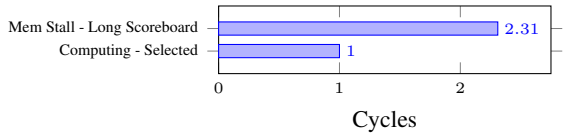
Secondly, we again conduct the warp state statistics analysis on the backward pass of the FlashKAT group-wise rational function.  From observing the warp state statistics in Figure \ref{fig:warp‐cycles‐eff}, we can see that only the ``Stall Long Scoreboard'' is still longer than the ``Selected'' compute state; all the other 8 memory stalls are shorter than compute.  Furthermore, the ``Stall Long Scoreboard'' has reduced from 981.51 cycles to 2.31 cycles. This result indicates the new kernel spends more time performing computations than waiting for memory transfers.

\subsection{Image Recognition Training Speed and Accuracy}
\begin{table}[h]
\centering
\small
\begin{tabular}{l l c c c}
\toprule
Model         & \#Param. & Top-1 & Train Thp. (images/s) \\
\midrule
ViT-T            & 5.7 M   &  72.7 & 8954.97 ($\pm$ 17.36) \\
DeiT-T                  & 5.7 M   &  72.2 & 8954.97 ($\pm$ 17.36) \\
KAT-T                   & 5.7 M   &  74.6 & 87.73 ($\pm$0.01) \\
FlashKAT-T               & 5.7 M   &  74.6 & 6317.90 ($\pm$2.65)\\
\midrule
ViT-S              & 22.1 M  &  78.8 & 5311.71 ($\pm$ 6.58)\\
DeiT-S                    & 22.1 M  &  79.8 & 5311.71 ($\pm$ 6.58) \\
KAT-S                    & 22.1 M  &  81.2 & 43.28 ($\pm$ 0.01)\\
FlashKAT-S               & 22.1 M  &  81.4 & 3741.91 ($\pm$ 0.08)\\
\midrule
ViT-B                & 86.6 M  &  79.1 & 2457.15 ($\pm$ 2.08)\\
DeiT-B                   & 86.6 M  &  81.8 & 2457.15 ($\pm$ 2.08) \\
KAT-B                  & 86.6 M  & 82.3 & 21.24 ($\pm$ 0.02)\\
FlashKAT-B             & 86.6 M  & 82.2 & 1801.75 ($\pm$ 0.683)\\
\bottomrule
\end{tabular}
\caption{Comparative analysis of model performance and computational efficiency on ImageNet-1K.}
\label{tab:imagenet_results}
\end{table}
We validate that FlashKAT's training capabilities are consistent with KAT by training it on ImageNet-1k \cite{russakovsky2015imagenet}. During training, we measure FlashKAT's training throughput to compare it with KAT, ViT, and DeiT. The data loader time is excluded from this measurement because it depends on external factors (e.g., CPU, main memory, etc.). We measure throughput in images/second as the average of 100 samples, and provide 95\% confidence intervals. 

Our training procedure follows the same curriculum as \cite{yang2024kolmogorov}, using the hyperparameters of DeiT \cite{touvron2021training}. As such, we train using a batch size of 1024 with the AdamW optimizer \cite{loshchilov2017decoupled}.  For the models, we apply Mimetic initialization to the attention layers \cite{trockman2023mimetic} and set the patch size to 16. In GR-KAN, each of the 8 groups shares its 6 numerator coefficients and has 4 unique denominator coefficients.  The first layer of GR-KAN has its group-wise rational function initialized to the identity function, and the second layer is initialized to a Swish function \cite{ramachandran2017searching}. For data augmentation and regularization techniques, we apply RandAugment \cite{cubuk2020randaugment}, Mixup \cite{zhang2017mixup}, CutMix \cite{yun2019cutmix}, Random Erasing \cite{zhong2020random}, weight decay, Label Smoothing \cite{szegedy2016rethinking} and Stochastic Depth \cite{huang2016deep}.  We conduct all experiments using an H200 GPU. The Appendix provides the complete set of model configurations and training hyperparameters.

We provide the image recognition accuracy of FlashKAT-T, FlashKAT-S, and FlashKAT-B models on the ImageNet-1k dataset in Table \ref{tab:imagenet_results}.  Importantly, FlashKAT-T, FlashKAT-S, and FlashKAT-B all match their original counterparts and continue to outperform their ViT and DeiT alternatives in image classification.  We also report the respective training throughput in Table \ref{tab:imagenet_results}.  Our results demonstrate that the proposed FlashKAT-T, FlashKAT-S, and FlashKAT-B are 72.0x, 86.5x, and 84.5x faster than their KAT counterparts. This significantly reduces the performance gap to ViT, making KAT-based models a viable option for vision tasks.

\subsection{Reduced Gradient Rounding Error}
\begin{table}[h]
  \centering
  \small
  \begin{tabular}{ccc}
    \toprule
    \multicolumn{3}{c}{\textbf{KAT}} \\
    \midrule
    Gradient & Mean Absolute Error                        & Variance        \\
    \midrule
    $\mathbf{dA}$       & $8.84\times10^{-2}\;(\pm3.40\times10^{-3})$  & $1.45\times10^{-2}$ \\
    $\mathbf{dB}$       & $9.63\times10^{-2}\;(\pm9.87\times10^{-3})$  & $8.11\times10^{-2}$ \\
    \midrule
    \multicolumn{3}{c}{\textbf{FlashKAT}} \\
    \midrule
    Gradient & Mean Absolute Error                         & Variance        \\
    \midrule
    $\mathbf{dA}$       & $8.42\times10^{-4}\;(\pm3.28\times10^{-5})$ & $1.35\times10^{-6}$ \\
    $\mathbf{dB}$       & $9.81\times10^{-4}\;(\pm1.15\times10^{-4})$ & $1.11\times10^{-5}$ \\
    \bottomrule
  \end{tabular}
    \caption{A comparison of mean absolute error (with 95\% confidence intervals) and variances.}
    \label{tab:comparison_grad}
\end{table}
In addition to a significant speedup, the proposed FlashKAT also reduces rounding errors when computing gradients for the group-wise rational function coefficients. This is a byproduct of our method, accumulating gradients in each block through a reduction sum rather than atomic adds. Table \ref{tab:comparison_grad} demonstrates that the mean absolute error in gradient outputs in FlashKAT is orders of magnitude smaller than in KAT.  As such, FlashKAT is less susceptible to rounding errors that may contribute toward unstable gradient updates (e.g., low-precision training). Due to space limitations, we provide more details in the Appendix.

\section{Conclusion}
\label{sec:conclusion}
Although KAN has achieved success, it has yet to see widespread adoption in large-scale NLP and computer vision tasks due to its increased parameter count, higher computational cost, lack of GPU hardware optimization, and training instability.  KAT has addressed many of KAN's shortcomings with the introduction of GR-KAN, but it still suffers from slow training. In this work, we conduct experiments to understand the source of the slowdown and identify that slow training speeds are a byproduct of memory stalls in the backward pass, a discovery that eluded prior work \cite{yang2024kolmogorov, molina2019pad, delfosse2020rationals, delfosse2021recurrent}. To address this bottleneck, we propose a novel restructuring of the group-wise rational function that allows efficient gradient accumulation by minimizing atomic add usage.  The resulting FlashKAT achieves an 86.5x speedup over KAT, while reducing gradient rounding errors.  The application of our work both enables wider adoption and expedites research on KAT.

\textbf{Limitations.} A main drawback preventing the use of KAN in the Transformer is KAN's increased computational cost.  With GR-KAN, KAT successfully reduced FLOPs, allowing it to nearly match the speed of a standard Transformer on the forward pass. Still, it remains orders of magnitude slower on the backward pass, making training computationally burdensome.  Our proposed FlashKAT, with a significant speedup, has largely closed the performance gap, but is still about 25\% slower than a standard Transformer. Even so, our work is an important step toward closing the gap between the KAT and standard Transformer in training speed, motivating future work to narrow it further.
\bibliography{aaai2026}
\input{appendix}

\end{document}

%% file: appendix.tex
\appendix
\section{Appendix}
\label{app:setup}
\subsection{Image Recognition Training Hyperparameters}
\label{app:hyper}
We provide the model architecture for KAT-T, KAT-S, and KAT-B in Table \ref{tab:size}.

\begin{table}[ht]
\centering

\begin{tabular}{lccccc}
\toprule
Model   & Layers & \shortstack[c]{Hidden\\size} & \shortstack[c]{MLP\\size} & Heads & Params  \\
\midrule
KAT-T  & 12     & 192             &  768     &   3   &  5.7 M \\
KAT-S & 12     & 384             & 1536     &   6   & 22.1 M \\
KAT-B  & 12     & 768             & 3072     &  12   & 86.6 M \\
\bottomrule
\end{tabular}
\caption{Details of KAT model variants.}
\label{tab:size}
\end{table}

We provide the training hyperparameters for KAT-T, KAT-S, and KAT-B in Table \ref{tab:hyper}.
\begin{table}[h]
\centering
\setlength{\tabcolsep}{1pt}
\begin{tabular}{lccc}
\toprule
Hyperparameter                  & KAT-T      & KAT-S     & KAT-B      \\
\midrule
Input resolution                  & 224×224   & 224×224   & 224×224   \\
Epochs                            & 300       & 300       & 300       \\
Batch size                        & 1024      & 1024      & 1024      \\
Optimizer                         & AdamW     & AdamW     & AdamW     \\
Adam $\epsilon$                   & $1\times10^{-8}$ & $1\times10^{-8}$ & $1\times10^{-8}$ \\
Adam $(\beta_1,\beta_2)$          & (0.9, 0.999) & (0.9, 0.999) & (0.9, 0.999) \\
Learning rate                     & $1\times10^{-3}$ & $1\times10^{-3}$ & $1\times10^{-3}$ \\
Learning rate decay               & Cosine    & Cosine    & Cosine    \\
Gradient clipping                 & None      & None      & None      \\
Warmup epochs                     & 5         & 5         & 5         \\
Weight decay                      & 0.05      & 0.05      & 0.05      \\
Rand Augment                      & 9/0.5     & 9/0.5     & 9/0.5     \\
\shortstack{Repeated \\augmentation}             & off       & off       & off       \\
CutMix                            & 1.0       & 1.0       & 1.0       \\
MixUp                             & 0.8       & 0.8       & 0.8       \\
\shortstack{CutMix–MixUp \\switch prob.}         & 0.5       & 0.5       & 0.5       \\
Random erasing prob.              & 0.25      & 0.25      & 0.25      \\
Label smoothing                   & 0.1       & 0.1       & 0.1       \\
\shortstack{Peak stochastic \\ depth rate }       & 0.1       & 0.1       & 0.4       \\
EMA decay rate                    & 0.9999    & 0.9999    & 0.9999    \\
\bottomrule
\end{tabular}
\caption{Hyperparameters of KAT on ImageNet-1K classification.}
\label{tab:hyper}
\end{table}  

\subsection{Reduced Gradient Rounding Error Expanded}
\label{app:grad}
FlashKAT also reduces rounding error when computing the gradients for the learned group-wise rational function coefficients, a byproduct of our implementation accumulating gradients in each block through sum reduction rather than using atomic adds. We validate that FlashKAT reduces gradient rounding errors by constructing an experiment measuring the coefficient gradient rounding errors.  Our experiment generates $\mathbf{X}\in\mathbb{R}^{1024\times197\times768}$, $\mathbf{dO}\in\mathbb{R}^{1024\times197\times768}$, $\mathbf{A}\in\mathbb{R}^{8\times6}$, and $\mathbf{B}\in\mathbb{R}^{8\times4}$ whose entries are distributed according to $\mathcal{N}(0,1)$.  Then we compute $\mathbf{dA}\in\mathbb{R}^{8\times6}$ and $\mathbf{dB}\in\mathbb{R}^{8\times4}$ using the KAT group-wise rational function backward pass method in float64 and float32, and the FlashKAT group-wise rational function backward pass method in float32.  We measure the mean absolute error (MAE) over 100 passes between the float64 gradient output and two float32 outputs.  All these experiments were done on an RTX 4060 Ti. We report the mean, 95\% confidence interval, and variance in Table \ref{tab:comparison_grad2}.

\begin{table}[h]
  \centering
  \small
  \begin{tabular}{ccc}
    \toprule
    \multicolumn{3}{c}{\textbf{KAT}} \\
    \midrule
    Gradient & Mean Absolute Error                        & Variance        \\
    \midrule
    $\mathbf{dA}$       & $8.84\times10^{-2}\;(\pm3.40\times10^{-3})$  & $1.45\times10^{-2}$ \\
    $\mathbf{dB}$       & $9.63\times10^{-2}\;(\pm9.87\times10^{-3})$  & $8.11\times10^{-2}$ \\
    \midrule
    \multicolumn{3}{c}{\textbf{FlashKAT}} \\
    \midrule
    Gradient & Mean Absolute Error                         & Variance        \\
    \midrule
    $\mathbf{dA}$       & $8.42\times10^{-4}\;(\pm3.28\times10^{-5})$ & $1.35\times10^{-6}$ \\
    $\mathbf{dB}$       & $9.81\times10^{-4}\;(\pm1.15\times10^{-4})$ & $1.11\times10^{-5}$ \\
    \bottomrule
  \end{tabular}
    \caption{A comparison of absolute differences in means (with 95\% confidence intervals) and variances.}
    \label{tab:comparison_grad2}
\end{table}

From Table \ref{tab:comparison_grad2}, we can see that the MAE between the KAT float64 and FlashKAT float32 gradient outputs is nearly 2 orders of magnitude smaller compared to the MAE with the KAT float32 gradient outputs.  Such a result verifies that our implementation is less susceptible to rounding errors.  Although not extremely visible in training at float32, where rounding errors are already minor, the reduction in rounding errors from FlashKAT could be helpful for low-precision training where gradient updates are more unstable.